\documentclass[lettersize,journal]{IEEEtran}
\usepackage{amsmath,amsfonts}
\usepackage{algorithmic}
\usepackage{algorithm}
\usepackage{array}
\usepackage[caption=false,font=normalsize,labelfont=sf,textfont=sf]{subfig}
\usepackage{textcomp}
\usepackage{stfloats}
\usepackage{url}
\usepackage{verbatim}
\usepackage{graphicx}
\usepackage{cite}
\usepackage{amssymb}
\usepackage{xcolor}
\usepackage{booktabs} 
\usepackage{multirow} 
\usepackage{pifont}
\newcommand{\etal}{\textit{et al.}}

\begin{document}

\title{DDTNet: Degradation Disentanglement and Transfer Network for Test-Time All-in-One De-weathering Adaptation}

\author{Kuan-Hung Lin*, Fu-Jen Tsai*, Yan-Tsung Peng,~\IEEEmembership{Senior Member,~IEEE,} \\
        Min-Hung Chen, Chia-Wen Lin,~\IEEEmembership{Fellow,~IEEE,} and Yen-Yu Lin,~\IEEEmembership{Senior Member,~IEEE}
\thanks{*Kuan-Hung Lin and Fu-Jen Tsai contributed equally to this work.}
\thanks{Kuan-Hung Lin and Yen-Yu Lin are with National Yang Ming Chiao Tung University, Hsinchu 300093, Taiwan (e-mail: guan.cs13@nycu.edu.tw; lin@cs.nycu.edu.tw).}%
\thanks{Fu-Jen Tsai and Chia-Wen Lin are with National Tsing Hua University, Hsinchu 300044, Taiwan (e-mail: fjtsai@gapp.nthu.edu.tw; cwlin@ee.nthu.edu.tw).}%
\thanks{Yan-Tsung Peng is with National Chengchi University, Taipei 116011, Taiwan (e-mail: ytpeng@cs.nccu.edu.tw).}%
\thanks{Min-Hung Chen is with NVIDIA (e-mail: minhungc@nvidia.com).}%
\thanks{This work has been submitted to IEEE Transactions on Multimedia.}
}

\markboth{}%
{Lin \MakeLowercase{\textit{et al.}}: DDTNet: Degradation Disentanglement and Transfer Network for Domain-Adaptive All-in-One Image De-weathering}


\maketitle

\begin{abstract}
All-in-one adverse weather image restoration aims to remove multiple degradations, such as rain, haze, and snow, using a single unified model. 
Despite their broad applicability, existing methods typically compromise performance, delivering balanced but suboptimal results for individual degradation types. 
This issue becomes more pronounced when a domain gap exists between training and testing data.
{\color{black} 
Based on the observation that modeling degradation patterns is more tractable than recovering clean content, we propose the Degradation Disentanglement and Transfer Network (DDTNet), a framework designed specifically for degradation transfer.}
By disentangling degradation patterns from target-domain degraded images and transferring them to source-domain clean images, DDTNet generates domain-adaptive paired training data. 
{\color{black} These pairs are then used to fine-tune restoration models, significantly enhancing their adaptability across diverse weather conditions and domains.}
The core of DDTNet is the Degradation Disentanglement Module (DDM), which comprises Degradation Coupled Attention (DCA) to capture both general and weather-specific features, thereby enabling effective disentanglement and transfer of degradation patterns.
Experimental results demonstrate that DDTNet significantly and consistently improves existing all-in-one models across real-world deraining, desnowing, and dehazing datasets.
\end{abstract}

\begin{IEEEkeywords}
Low-level vision, adverse weather image restoration, domain adaptation 
\end{IEEEkeywords}

\begin{figure*}[t]
  \centering
  \includegraphics[width=\textwidth]{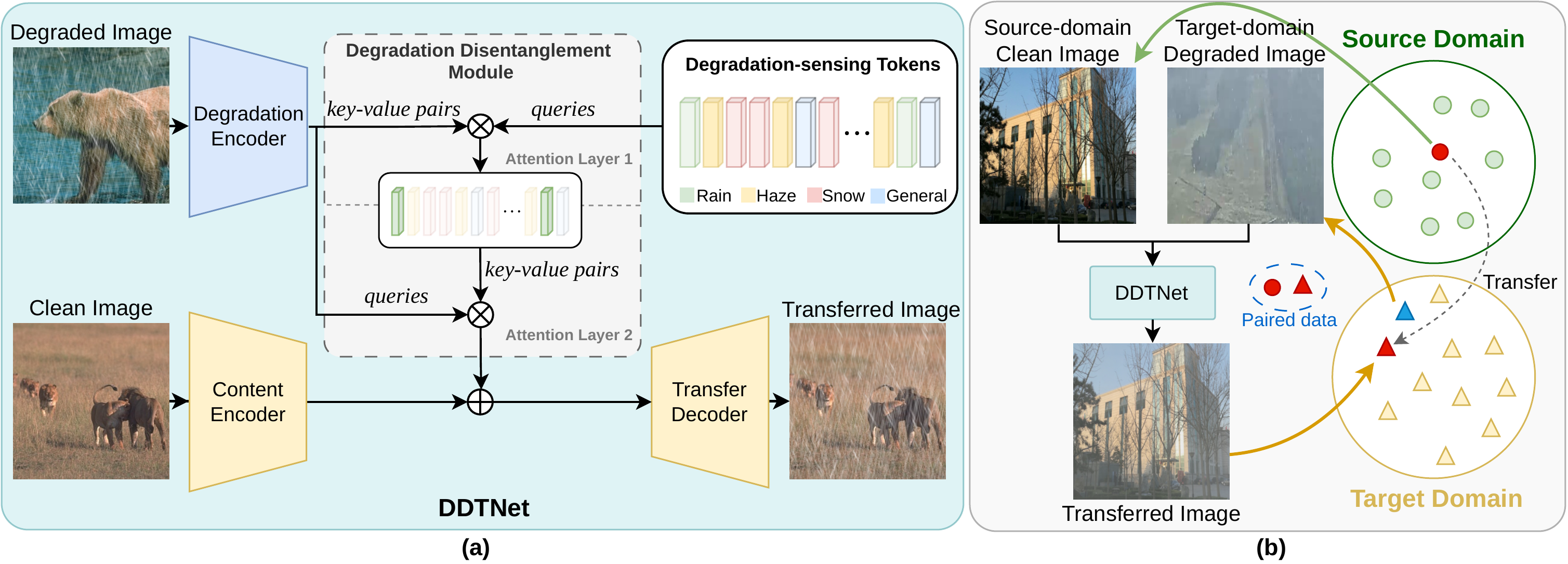}
  \vspace{-0.0in}
  \caption{%
  \textbf{DDTNet for all-in-one model adaptation.}
  {\color{black}(a) DDTNet is a dual-branch encoder-decoder architecture with a Degradation Disentanglement Module (DDM). A degraded image is first fed into the degradation encoder branch, followed by a two-stage attention mechanism guided by Degradation-sensing Tokens encoding diverse weather cues, to disentangle the underlying degradation patterns. These disentangled degradation patterns are then transferred to a clean image processed by the content encoder branch and finally fused by the decoder to synthesize the target degraded image.} 
  (b) For test-time adaptation, DDTNet disentangles degradation patterns from a target-domain degraded image and transfers them to a source-domain clean image, generating paired images for fine-tuning an arbitrary all-in-one restoration model.
      }
  \label{fig:teaser}
  \vspace{-0in}
\end{figure*}

\section{Introduction}
Image de-weathering aims to recover clean, high-quality images from inputs degraded by atmospheric conditions. 
In the past, research has focused on specialized solutions for individual weather conditions, such as deraining~\cite{Jiang_2022_ACMMM,Hu_2019_CVPR,Wang_2020_CVPR,Wang_2019_CVPR,8099669,Kui_2020_CVPR,Li_2019_CVPR,NeRD-Rain,gao2024efficient,10415525,9388918}, dehazing~\cite{Wu_2021_CVPR,guo2022dehamer,song2023vision,9010659,Hardgan,FSDGN,qin2020ffa,Zhang_2024_CVPR,fang2025guided,9726872,10292884}, and desnowing~\cite{JSTASRChen,chen2021all,liu2018desnownet,zhang2021deep,7934435}.  
However, because weather conditions are inherently unpredictable and vary over time, de-weathering methods designed for a single type of weather-induced degradation have limited practical applicability.

To overcome these limitations, all-in-one image de-weathering~\cite{AirNet,Valanarasu_2022_CVPR,Chen2022MultiWeatherRemoval,Park_2023_CVPR,potlapalli2023promptir,cui2025adair,sun2024restoring,zhang2023ingredient,9966834, 10855564, 10480609} has recently emerged as a significant research direction. Unlike task-specific approaches, a unified framework aims to handle multiple weather scenarios, making it more scalable and adaptable to real-world applications.
To this end, several studies have explored jointly learning degradation-specific and degradation-general representations within a shared architecture~\cite{Valanarasu_2022_CVPR, Park_2023_CVPR, potlapalli2023promptir, tian2025degradation}.

Despite their specific designs and advancements, existing all-in-one methods face two critical bottlenecks: multi-task trade-off and domain gaps.
First, the joint multi-task training strategy inherently trades off performance across tasks. Although it enables a single model to handle diverse degradations, it usually yields suboptimal results for single tasks compared to specialized single-task models. 
{\color{black}
Second, while most image restoration methods are sensitive to domain gaps between training and testing data, all-in-one methods are even more vulnerable as they capture a wide spectrum of degradation statistics within a single architecture. Consequently, these domain shifts severely hinder all-in-one methods' generalization and effectiveness in real-world deployment.}

{\color{black}
To address these issues, previous studies~\cite{Valanarasu_2022_CVPR,Chen2022MultiWeatherRemoval,Yang_2024_CVPR,10473168} have focused on improving restoration performance by refining restoration models. 
However, recovering unknown clean content directly from complex degradations is a highly ill-posed problem. 
In this work, we pivot from {\em content recovery} to {\em degradation modeling} based on a key observation: modeling degradation patterns is inherently more tractable than directly recovering clean images, as these degradation patterns, like raindrops or haze, are relatively uniform, smooth, and dominated by low-frequency structures (Empirical evidence supporting this observation is provided in Section~\ref{sec:exp}.B.).}

{\color{black}
Consequently, we propose the Degradation Disentanglement and Transfer Network (DDTNet), which focuses on degradation transfer to enhance the performance of a restoration model.}
%
It disentangles degradation patterns from target-domain weather-degraded images and transfers them onto source-domain clean images, acting as targeted data augmentation. 
{\color{black}
This process generates domain-adaptive pairs that reflect the target distribution and fine-tune all-in-one restoration models, thereby bridging the domain gap and improving restoration quality.}
%

As shown in Figure~\ref{fig:teaser}(a), the technical core of DDTNet is the Degradation Disentanglement Module (DDM), which uses a Degradation Coupled Attention (DCA) mechanism to effectively disentangle diverse degradation patterns.  
The DCA is a two-stage attention mechanism applied to learnable {\em degradation-sensing tokens}.
In the first stage, these tokens serve as queries to retrieve degraded features from the input image, yielding {\em degradation tokens} that encode the underlying degradation patterns. 
In the second stage, the roles are reversed: the image tokens act as queries to aggregate information stored in these degradation tokens for distilling degradation features.
Repeating this two-stage interaction can progressively disentangle degradation patterns from the image.

This disentanglement strategy is the core of our work, which allows DDTNet to transfer isolated degradation patterns from a target-domain degraded image to a source-domain clean image.
As shown in Figure~\ref{fig:teaser}(b), this transfer process generates domain-adaptive paired data, consisting of clean source images with target-domain degradation patterns.
These synthetic yet aligned pairs not only embed target-domain characteristics but also adapt to weather-specific conditions such as rain, haze, snow, or even mixed weather, which are then used to fine-tune de-weathering models. 
Through explicitly aligning the restoration models with target-domain degradations, DDTNet improves model performance and restoration quality in challenging real-world scenarios.  

This work makes three primary contributions.
First, we present DDTNet to tackle two critical challenges in all-in-one image restoration: suboptimal cross-task performance and sensitivity to domain gaps. 
DDTNet disentangles and transfers degradation patterns from target-domain degraded images to source-domain clean images, hence generating paired, domain-adaptive images for fine-tuning all-in-one restoration models.
Second, we propose DCA, a two-stage attention mechanism that effectively identifies and separates degradation patterns from degraded images by alternating the roles of queries and key-value pairs between degradation sensing tokens and image tokens.
Finally, we demonstrate through extensive experiments that DDTNet significantly improves existing all-in-one restoration models on real-world deraining, desnowing, and dehazing benchmarks.

\section{Related Work}

This section reviews two research topics relevant to the development of our method, including all-in-one image restoration and domain adaptation for image restoration.

\subsection{All-in-One Image Restoration}
All-in-one image restoration aims to handle multiple degradations, such as rain, haze, and snow, using a unified model. 
To tackle this task, several studies~\cite{Chen2022MultiWeatherRemoval, zhu2023Weather,potlapalli2023promptir,cui2025adair,tian2025degradation, 9966834} explore learning both degradation-specific and degradation-agnostic features in the network.
Potlapalli~\etal~\cite{potlapalli2023promptir} integrated degradation-specific cues into a unified model via learnable prompts to handle diverse degradations.
Cui~\etal~\cite{cui2025adair} extracted degradation-specific frequency subbands to adaptively address different degradations through frequency mining and modulation. 
Furthermore, Tian~\etal~\cite{tian2025degradation} introduced degradation-aware feature perturbations to align degradation-specific prompts within a unified architecture.

Although these methods demonstrate the potential of unified networks to handle diverse degradations, they often yield suboptimal performance on individual tasks due to the inherent interference in multi-task joint training.
Moreover, the domain gap between training and testing data frequently causes significant performance drops, severely degrading restoration results in real-world scenarios.

{\color{black}
To overcome these inherent limitations, our proposed DDTNet introduces a distinctly different test-time adaptation strategy. Rather than developing a complex unified restoration architecture, DDTNet synthesizes realistic training pairs by disentangling degradation patterns from the unseen target domain and transferring them to clean reference images. 
By leveraging these synthesized pairs for test-time fine-tuning, DDTNet not only narrows the domain gap but also allows a pre-trained all-in-one model to concentrate on the current degradation patterns, thereby effectively alleviating the suboptimal performance issues associated with joint multi-task training.
}

\subsection{Domain Adaptation for Image Restoration}

Domain adaptation in the field of image restoration seeks to narrow the domain gap between the source and target domains, ensuring that models trained on synthetic data generalize well to real-world scenarios.
Early research primarily focuses on single-degradation scenarios~\cite{Wei2021DerainCycleGAN,DADehaze}. 
For instance, Wei~\etal~\cite{Wei2021DerainCycleGAN} and Shao~\etal~\cite{DADehaze} utilized CycleGAN~\cite{CycleGAN2017} to generate pseudo-training data for deraining and dehazing tasks.
Patil~\etal~\cite{Patil_2023_ICCV} proposed a domain translation approach that converts degraded images across weather types to learn weather-invariant features via generative adversarial networks (GANs). However, GAN-based translation often suffers from uncontrollable results, which limits its reliability for consistent weather modeling and synthesis.
More recently, Chi~\etal~\cite{MetaDeblur} introduced a meta-auxiliary learning strategy to enable fast test-time adaptation for deblurring.

Although these methods alleviate the domain shift issue in single-degradation scenarios, they lack the flexibility to generalize across diverse degradation types. 
To address this, Liao~\etal~\cite{liao2024denoising} proposed a general domain adaptation framework built upon a pre-trained diffusion model~\cite{NEURIPS2020_4c5bcfec}, which computes a diffusion loss to align the distributions of restored synthetic and real-world images.  
However, since this method is not tailored for all-in-one image restoration, which requires both degradation-specific and degradation-general features, it cannot effectively handle multiple degradations within a unified framework. 

{\color{black}
The most closely related work to our method is our early study, the weather-transfer network (WTNet)~\cite{Huang_2025_BMVC, Dark_Channel_Prior}, which transfers target-domain weather patterns to clean images.
However, WTNet relies heavily on the physical priors of the atmospheric scattering model (ASM)~\cite{Fattal_2008, Dark_Channel_Prior} to disentangle and decompose weather patterns into two parameters, atmospheric light and haze density, which is insufficient to address diverse real-world degradation patterns, as real-world scenes often contain mixed and spatially varying degradations beyond the assumptions of ASM. 
In contrast, the proposed DDTNet bypasses these rigid physical priors and adopts a more general approach by disentangling and transferring both degradation-specific and degradation-general features.
This enables DDTNet to transfer diverse degradation patterns from target domains to source-domain clean images, generating realistic training pairs for fine-tuning restoration models and improving restoration performance.
}

\begin{figure*}[t]
  \centering
  \includegraphics[width=\textwidth]{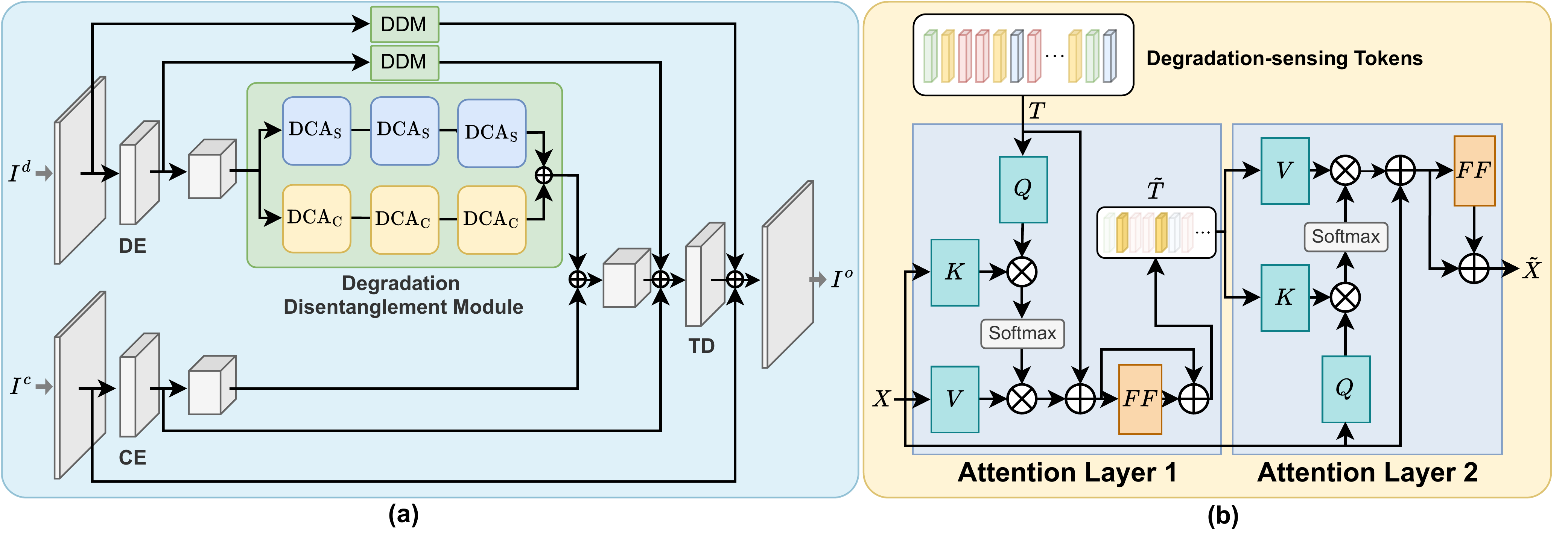}
  \vspace{-0.15in}
  \caption{%
  \textbf{Network architecture of DDTNet.}
  (a) DDTNet employs a dual-branch structure: the top branch extracts degradation features from a degraded image $I^{d}$ via a Degradation Encoder (DE) and a Degradation Disentanglement Module (DDM). The DDM leverages both spatial-wise and channel-wise Degradation Coupled Attention ($\mathrm{DCA_S}$ and $\mathrm{DCA_C}$) to isolate degradation features from image content. The bottom branch extracts content features from a clean image $I^{c}$ using a Content Encoder (CE). These dual-branch features are fused through a Transfer Decode (TD) to generate the degradation-transferred image $I^{o}$. 
  (b) DCA is a two-stage attention mechanism: learnable degradation-sensing tokens $T$ first retrieve degradation patterns from a degraded image to yield degradation tokens $\tilde{T}$, which are subsequently queried by image tokens to isolate degradation patterns from scene content by distilling only degradation features. {\color{black}Finally, a feedforward (FF) layer is applied to further refine these extracted representations.}}  
  \label{fig:model_arc}
  \vspace{-0.13in}
\end{figure*}

\section{Proposed Method}
{\color{black}
In this section, we present the proposed Degradation Disentanglement and Transfer Network (DDTNet), a framework specifically designed to address two primary challenges in all-in-one image restoration: 1) the suboptimal task-specific performance of an all-in-one model on individual tasks compared to specialized single-task models and 2) the domain gap between synthetic training data and complex real-world scenes. 
Unlike previous methods that rely on task-specific physical priors, DDTNet employs a generalized degradation-transfer-driven strategy to synthesize high-quality, domain-adaptive training pairs. 

In the following, we first provide an architectural overview of DDTNet. We then detail the core design of the Degradation Disentanglement Module (DDM) and its Degradation-Coupled Attention (DCA) mechanism, which effectively disentangles and transfers diverse degradation patterns. Finally, we elaborate on the complete test-time domain adaptation pipeline utilizing our generated adaptive pairs.
}

\subsection{Overview}  
{\color{black}
DDTNet is designed to transfer diverse degradation patterns from a degraded image in an unseen target domain to a clean image in the source domain. By doing so, the clean source image and its degradation-transferred counterpart form a domain-adaptive training pair.}
Consequently, DDTNet facilitates test-time adaptation (TTA) through the generation of these pairs, enabling model fine-tuning and significantly improving restoration performance under unseen degradation conditions.

As illustrated in Figure~\ref{fig:model_arc}(a), DDTNet adopts a dual-branch architecture: the top branch for extracting degradation features and the bottom branch for compiling content feature representation.
{\color{black}
Given a degraded image $I^d \in \mathbb{R}^{H \times W \times 3}$ and a clean image $I^c \in \mathbb{R}^{H \times W \times 3}$, DDTNet employs a Degradation Encoder (DE) and a Content Encoder (CE) for feature extraction.  
Both encoders consist of three scales, with each scale containing residual blocks. 
The two encoders produce multi-scale degradation features $\mathrm{DE}(I^d) = \{F_i^{d} \in \mathbb{R}^{\frac{H}{2^{i}} \times \frac{W}{2^{i}} \times C_i}\}$ and content features $\mathrm{CE}(I^c) = \{F_i^{c} \in \mathbb{R}^{\frac{H}{2^{i}} \times \frac{W}{2^{i}} \times C_i}\}$, respectively, where $i \in \{0,1,2\}$ indexes the feature scale and $C_i = 32 \cdot 2^i$. 
}

At each scale $i$, the degradation features $F_i^{d}$ are processed by the proposed Degradation Disentanglement Module (DDM). The DDM integrates parallel Spatial-wise Degradation-Coupled Attention ($\mathrm{DCA_S}$) and Channel-wise Degradation-Coupled Attention ($\mathrm{DCA_C}$) layers to distill degradation cues while suppressing scene content. 
Specifically, the DDM produces the distilled degradation features $\tilde{F}_i^{d} = \mathrm{DDM}(F^{d}_{i}) \in \mathbb{R}^{\frac{H}{2^{i}} \times \frac{W}{2^{i}} \times C_i}$ via
{%
\setlength{\abovedisplayskip}{3pt}%
\setlength{\belowdisplayskip}{3pt}%
\begin{equation}
\begin{aligned}
\mathrm{DDM}(F^{d}_{i})
&= \mathrm{Conv}\!\Big(\mathrm{Concat}\!\big(
(\mathrm{DCA_S})^{3}(F^{d}_{i}),\\[-2pt]
&\qquad\quad
(\mathrm{DCA_C})^{3}(F^{d}_{i})
\big)\Big),
\end{aligned}
\end{equation}
}%
where $(\mathrm{DCA_S})^3$ and $(\mathrm{DCA_C})^3$ denote the sequential application of three spatial and three channel attention layers, respectively.
{\color{black}
The specific mechanisms of $\mathrm{DCA_S}$ and $\mathrm{DCA_C}$ for simultaneous degradation cue distillation and scene content suppression are detailed in the next subsection.}

{\color{black}
Subsequently, we apply the Transfer Decoder (TD), a network containing three scales with channel sizes of $128$, $64$, and $32$, respectively, to fuse multi-scale distilled degraded features $\{\tilde{F}^{d}_{i}\}$ and content features $\{F^{c}_{i}\}$, for generating the degradation-transferred image $I^o \in \mathbb{R}^{H \times W \times 3}$ via
}
\begin{equation}
F^i = \mathrm{TD}_i\left(\mathrm{Concat}\big(\tilde{F}^{d}_{i}, F^{c}_{i}, F^{(i+1)}\uparrow\big)\right),
\mbox{ for } i \in \{0,1,2\},
\end{equation}
where $F^3:=\emptyset$, and $I^o := F^0$. Here, operator $(\cdot) {\uparrow}$ denotes a $2\times$ upsampling operator, and $\{\text{TD}_i\}_{i=0}^2$ are the Transfer Decoder for different scales. 


\subsection{Degradation Coupled Attention (DCA)}

Our proposed DCA is specifically designed to disentangle degradation patterns not only in single-degradation scenarios but also under complex mixed degradations, such as rain combined with haze or snow combined with haze. 
As illustrated in Figure~\ref{fig:model_arc}(b), DCA adopts a two-stage attention mechanism guided by degradation-sensing tokens, enabling it to capture both degradation-general and degradation-specific features.

Let the input features of DCA be $X \in \mathbb{R}^{N \times D}$, representing $N$ tokens of dimension $D$. 
In the first stage, $X$ is processed by linear projection layers $\mathrm{L}$ to generate the key and value features, i.e., $X^k \in \mathbb{R}^{N \times D}$ and $X^v \in \mathbb{R}^{N \times D}$, which is formulated as 
\begin{equation}
    (X^k, X^v) = \mathrm{L}(X).
\label{eq:LN_Layer1}
\end{equation}

We then introduce a set of learnable {\em degradation-sensing tokens} $T \in \mathbb{R}^{M \times D}$, consisting of $M$ tokens of dimension $D$. These tokens are linearly projected through $\mathrm{L'}$ to serve as queries, $T^q = \mathrm{L'}(T)$, and can retrieve degradation-related information from $X$ via cross-attention, yielding the {\em degradation tokens} $\tilde{T} \in \mathbb{R}^{M \times D}$:
\begin{equation}
    \tilde{T} = \mathrm{softmax}(\frac{T^q  (X^k)^{\top}}{\sqrt{D}})  X^v + T,
\label{eq:attn_Layer1}
\end{equation} 
where $\top$ denotes the transpose operation. In practice, we set $M=256$. The resulting $\tilde{T}$ can be regarded as a set of degradation kernels that are dynamically generated through the interaction between the input degraded image features and the degradation-sensing tokens.

In the second stage, the roles are reversed: the input features $X$ act as queries to aggregate the information stored in the degradation tokens $\tilde{T}$, thereby distilling degradation features. 
Specifically, we process $X$ and $\tilde{T}$ through linear projection layers $\mathrm{U}$ and $\mathrm{U'}$ to generate the query, key, and value features, denoted as $X^q \in \mathbb{R}^{N \times D}$, $\tilde{T}^k \in \mathbb{R}^{M \times D}$, and $\tilde{T}^v \in \mathbb{R}^{M \times D}$, respectively:
\begin{equation}
    X^q = \mathrm{U}(X) 
    \mbox{ and }
    (\tilde{T}^k, \tilde{T}^v) = \mathrm{U'}(\tilde{T}). 
\label{eq:LN_Layer2}
\end{equation} 
It follows that the distilled degradation features $\tilde{X} \in \mathbb{R}^{N \times D}$, which retain degradation cues while suppressing scene content, are obtained via
\begin{equation}
    \tilde{X} = \mathrm{softmax}(\frac{X^q (\tilde{T}^k)^{\top}}{\sqrt{D}}) \tilde{T}^v + X.
\label{eq:attn_Layer2}
\end{equation}

To effectively disentangle degradation patterns in the proposed DDM, we apply DCA along both the spatial and channel dimensions of the features. 
The spatial-wise DCA ($\mathrm{DCA_S}$) emphasizes the geometric structure and spatial distribution of degradations, while the channel-wise DCA ($\mathrm{DCA_C}$) concentrates on the contrast and intensity attenuation of degradation, as detailed below.

\paragraph{Spatial-wise Degradation Coupled Attention ($\mathrm{DCA_S}$)}
Let the input to $\mathrm{DCA_S}$ be $X \in \mathbb{R}^{H \times W \times C}$. 
We first flatten $X$ into a 2D tensor $X \in \mathbb{R}^{HW \times C}$, treating it as $HW$ tokens of $C$ dimensions. 
The reshaped input is then processed by the DCA operations as defined in Equations~(\ref{eq:LN_Layer1})--(\ref{eq:attn_Layer2}). 
Finally, the resulting features are projected back to the original spatial dimensions $H \times W \times C$ to yield the output of $\mathrm{DCA_S}$.

\paragraph{Channel-wise Degradation Coupled Attention ($\mathrm{DCA_C}$)}
Let the input feature map to $\mathrm{DCA_C}$ be $X \in \mathbb{R}^{H \times W \times C}$. 
We first perform spatial pooling to $X$ to obtain a downsampled representation $X_{\text{pool}} \in \mathbb{R}^{24 \times 24 \times C}$, and then reshape it into a 2D tensor of size $\mathbb{R}^{C \times 576}$, corresponding to $C$ tokens of $576$ dimensions.
The 2D tensor is then processed by the DCA operations as defined in Equations~(\ref{eq:LN_Layer1})--(\ref{eq:attn_Layer2}), and the output is reshaped back to $24 \times 24 \times C$.
Finally, bilinear upsampling is utilized to restore the resolution to $H \times W \times C$, yielding the output of $\mathrm{DCA_C}$.

\subsection{Loss Function}
DDTNet is trained on degraded–clean-mixed triplets $\{(I^d_i,\, I^c_i,\, I^{m}_i)\}_{i=1}^{K}$, where the ground-truth mixed image $I^{m}_i$ preserves the scene content of $I^c_i$ while exhibiting the same degradation patterns as $I^d_i$, as described in Section \ref{details}. 
To supervise the generation of each degradation-transferred image $I^o_i = \mathrm{DDTNet}(I^d_i, I^c_i)$, we adopt the $\ell_1$ reconstruction loss, defined as 
\begin{equation}
\mathcal{L} = \frac{1}{K} \sum_{i=1}^K \lVert I^o_i - I^{m}_i \rVert_1.
\end{equation}

\subsection{Domain-Adaptive Fine-Tuning Process}
After training, DDTNet is employed to transfer degradation patterns from target-domain degraded images \(\{{I}^{d}_{i}\}_{i=1}^{K}\) to source-domain clean images \(\{{I}^{c}_{i}\}_{i=1}^{K}\), randomly sampled from a source-domain dataset, {\color{black}where $K$ denotes the total number of images in the target dataset. }
As illustrated in Figure~\ref{fig:model_arc}(b), for each pair \(({I}^{d}_{i},{I}^{c}_{i})\), DDTNet generates a degradation-transferred image \({I}^{o}_{i}=\mathrm{DDTNet}({I}^{d}_{i},{I}^{c}_{i})\), which preserves the scene content of ${I}^{c}_{i}$ while embedding the degradation patterns from ${I}^{d}_{i}$. 
This process allows us to construct domain-adaptive training pairs $\mathcal{D}_{\mathrm{adapt}} = \{({I}^{o}_{i},{I}^{c}_{i})\}_{i=1}^K$, which are then used to update restoration models during testing.
%
By generating the aligned training pairs with target-domain degradations and their corresponding clean images, DDTNet effectively adapts restoration models to the target domain, therefore improving their generalization and performance.

\section{Experiments}
\label{sec:exp}

{\color{black}
In this section, we conduct extensive experiments to evaluate the effectiveness of the proposed DDTNet. We first describe our experimental setup and implementation details, followed by quantitative and qualitative comparisons with state-of-the-art methods and detailed ablation studies to analyze the contributions of individual components.
}

\subsection{Implementation Details}
\label{details}

{\color{black}
\paragraph{Datasets}
We construct a set of degraded-clean-mixed triplets $\{(I^d_i,\, I^c_i,\, I^{m}_i)\}_{i=1}^{K}$. Here, $I^d_i$ denotes a degraded image from the target domain, $I^c_i$ represents a clean image with different scene content, and $I^{m}_i$ is a synthesized degraded image that preserves the content of $I^c_i$ while incorporating the specific degradation patterns of $I^d_i$. 
{\color{black}
This strategy enables the construction of triplet data that share consistent degradation patterns but differ in scene content to train DDTNet.
}


The data synthesis process for generating the degraded image $I^m$ across different weather conditions is described as follows.
For rainy and snowy scenarios, we employ degradation masks from the Rain100H~\cite{Yang_2017_CVPR}, Rain100L~\cite{Yang_2017_CVPR}, and Snow100K~\cite{liu2018desnownet} datasets. Given a clean image $I^c \in \mathbb{R}^{H \times W \times 3}$, we extract a binary rain or snow mask $M \in \{0,1\}^{H \times W \times 1}$ from the corresponding degraded image $I^d$. We then synthesize the degraded image $I^m \in \mathbb{R}^{H \times W \times 3}$ as follows:
\begin{equation}
I^{m} \;=\; (1-\lambda M)\odot I^{c} \;+\; (\lambda M)\odot \mathbf{c},
\end{equation}
where $\odot$ denotes element-wise multiplication, $\lambda \in [0,1]$ is the mask coefficient controlling the degradation intensity, and $\mathbf{c} \in [0,1]^{1 \times 1 \times 3}$ represents the chromatic aberration value.
For hazy images, we leverage the RESIDE dataset~\cite{li2019benchmarking}, which provides the parameters to synthesize images sharing the same haze characteristics (i.e., haze density and atmospheric light) but differing in content. {\color{black}Following previous methods~\cite{Huang_2025_BMVC, Dark_Channel_Prior}}, we generate the hazy image $I^m \in \mathbb{R}^{H \times W \times 3}$ using the clean image $I^c$ with the specific haze parameters:
\begin{equation}
I^{m} = I^{c} \times {M}^t + \alpha \times (1-{M}^t),
\end{equation}
where $\alpha \in \mathbb{R}^{3}$ represents the atmospheric light, and $M^t = e^{-\beta \times d}$ denotes the transmission map, which is determined by the haze density $\beta \in \mathbb{R}^{1}$. 
}

During training, we sample $5,000$ triplets for each of the three tasks, yielding a total of $15,000$ pairs for jointly optimizing DDTNet and the restoration models. To evaluate the effectiveness of DDTNet, we use the real-world WeatherStream dataset~\cite{zhang2023weatherstream}, which contains $4,500$ hazy, $3,000$ rainy, and $3,960$ snowy images with corresponding clean counterparts. This dataset poses a particularly challenging benchmark as it includes not only single-degraded cases but also mixed degradations, such as rain with haze and snow with haze. 
%
Furthermore, we adopt the latest real-world dataset, WeatherBench~\cite{guan2025weatherbench}, which includes $200$ images for each of the three weather types (rain, haze, and snow) to evaluate the effectiveness of DDTNet.

\paragraph{DDTNet Configuration} We optimize DDTNet using the Adam optimizer with a learning rate of $1 \times 10^{-4}$, a batch size of $4$, and $150$ training epochs.
DDTNet comprises $31$ million parameters and achieves an inference time of $33$ ms on an NVIDIA RTX A5000 GPU for an input size of $256{\times}256$.

\paragraph{Restoration Models} 
To evaluate the effectiveness of DDTNet for domain adaptation, we adopt three state-of-the-art (SOTA) restoration models: PromptIR~\cite{potlapalli2023promptir}, AdaIR~\cite{cui2025adair}, and DFPIR~\cite{tian2025degradation}.
All models are initially trained in an all-in-one manner on the $15,000$ synthesized training pairs using their respective default training configurations.
Although the fine-tuning time naturally varies with model architecture and the volume of generated domain-adaptive pairs, we find that {\em a single epoch} of fine-tuning on our generated paired images is sufficient to achieve consistent performance gains. This demonstrates that DDTNet enables highly efficient domain adaptation without requiring extensive retraining.
%
%

\begin{table}[!t]
\small
\centering
\setlength{\tabcolsep}{2mm}
\caption{
\textbf{Hypothesis verification:} Comparison between clean image recovery and degradation pattern prediction across different degradation types (PSNR).
}
\vspace{-0in}
\resizebox{\linewidth}{!}{
\begin{tabular}{l|cccc}
\noalign{\hrule height 1.0pt}
Prediction Target & Rain & Snow & Haze & Average \\
\noalign{\hrule height 0.5pt}
Image Recovery & 32.83 & 30.51 & 38.67 & 34.15 \\
Degradation Prediction & \bf 34.63 & \bf 31.67 & \bf 41.21 & \bf 35.84 \\
\noalign{\hrule height 1.0pt}
\end{tabular}
}
\label{tab:mask_vs_gt}
\vspace{-0in}
\end{table}

\subsection{Why Degradation Pattern Prediction is Easier?}
{\color{black}
To validate our observation that predicting degradation patterns from a degraded image is more tractable than directly recovering its clean image content, we conduct a comparative experiment based on a standard U-Net architecture.
Table~\ref{tab:mask_vs_gt} compares the PSNR results of two prediction targets across three degradation types.
The first setting directly reconstructs clean images from degraded inputs, whereas the second predicts the corresponding degradation components, {\em i.e.}, rain masks from rainy images, snow masks from snowy images, and transmission maps from hazy images.
As shown in Table~\ref{tab:mask_vs_gt}, degradation pattern prediction achieves consistently higher PSNR values than clean image recovery for all three degradation types. These results support our hypothesis that degradation patterns are easier to model than clean image content, thereby motivating our degradation-transfer-based design.

}

\begin{table*}[t]
\centering
\begingroup
\setlength{\tabcolsep}{3pt}
\renewcommand{\arraystretch}{1.08}

\newcommand{\gaincolor}{black!65}
\newcommand{\gainfontsize}{\scriptsize}

\newcommand{\valgain}[2]{%
  \begin{tabular}{@{}c@{}}%
    #1 \\[-3pt]%
    {\gainfontsize\color{\gaincolor}(#2)}%
  \end{tabular}%
}

\newcommand{\bestvalgain}[2]{%
  \begin{tabular}{@{}c@{}}%
    \bfseries #1 \\[-3pt]%
    {\gainfontsize\color{\gaincolor}(#2)}%
  \end{tabular}%
}

\caption{
\textbf{Quantitative comparison of different test-time domain adaptation methods (StyleSSP, WTNet, and the proposed DDTNet) integrated with three baseline models (PromptIR, AdaIR, and DFPIR).} 
Evaluations are performed on the WeatherStream and WeatherBench datasets across three real-world weather types: rain, snow, and haze. The best results among the integrated methods for each baseline are highlighted in bold.}
\vspace{-0in}
\label{table:restoration_two}

\resizebox{\textwidth}{!}{%
\begin{tabular}{@{} c c 
| c c | c c | c c | c c 
|| c c | c c | c c | c c @{}}
\toprule
& & \multicolumn{8}{c||}{\textbf{WeatherStream~\cite{zhang2023weatherstream}}} & \multicolumn{8}{c}{\textbf{WeatherBench~\cite{guan2025weatherbench}}} \\
\cmidrule(lr){3-10}\cmidrule(l){11-18}
\multicolumn{2}{c|}{\textbf{Method}} 
& \multicolumn{2}{c|}{Rain} & \multicolumn{2}{c|}{Snow} & \multicolumn{2}{c|}{Haze} & \multicolumn{2}{c||}{Average} 
& \multicolumn{2}{c|}{Rain} & \multicolumn{2}{c|}{Snow} & \multicolumn{2}{c|}{Haze} & \multicolumn{2}{c}{Average} \\
\multicolumn{2}{c|}{} 
& PSNR$\uparrow$ & SSIM$\uparrow$ & PSNR$\uparrow$ & SSIM$\uparrow$ 
& PSNR$\uparrow$ & SSIM$\uparrow$ & PSNR$\uparrow$ & SSIM$\uparrow$ 
& PSNR$\uparrow$ & SSIM$\uparrow$ & PSNR$\uparrow$ & SSIM$\uparrow$ 
& PSNR$\uparrow$ & SSIM$\uparrow$ & PSNR$\uparrow$ & SSIM$\uparrow$ \\
\midrule

\multirow[c]{4}{*}[-14pt]{\bfseries PromptIR} 
& Baseline & 22.96 & 0.7620 & 21.36 & 0.7381 & 19.08 & 0.6894 & 21.13 & 0.7298
& 27.49 & 0.8327 & 22.52 & 0.7975 & 13.58 & 0.5783 & 21.20 & 0.7362 \\[4pt]
& +StyleSSP & \valgain{16.44}{-6.52} & \valgain{0.6890}{-0.0730} & \valgain{18.74}{-2.62} & \valgain{0.7046}{-0.0335} & \valgain{16.09}{-2.99} & \valgain{0.6682}{-0.0212} & \valgain{17.09}{-4.04} & \valgain{0.6873}{-0.0425} 
& \valgain{17.85}{-9.64} & \valgain{0.7010}{-0.1317} & \valgain{18.09}{-4.43} & \valgain{0.7174}{-0.0801} & \valgain{13.31}{-0.27} & \bestvalgain{0.5885}{+0.0102} & \valgain{16.42}{-4.78} & \valgain{0.6689}{-0.0673} \\[4pt]
& +WTNet & \valgain{23.24}{+0.28} & \valgain{0.7365}{-0.0255} & \valgain{21.90}{+0.54} & \valgain{0.7260}{-0.0121} & \valgain{19.75}{+0.67} & \valgain{0.6485}{-0.0409} & \valgain{21.62}{+0.49} & \valgain{0.7043}{-0.0255} 
& \valgain{25.61}{-1.88} & \valgain{0.8211}{-0.0116} & \valgain{21.95}{-0.57} & \valgain{0.7762}{-0.0213} & \valgain{13.65}{+0.07} & \valgain{0.5850}{+0.0067} & \valgain{20.41}{-0.79} & \valgain{0.7274}{+0.0088} \\[4pt]
& +DDTNet & \bestvalgain{23.82}{+0.86} & \bestvalgain{0.7699}{+0.0079} & \bestvalgain{22.12}{+0.76} & \bestvalgain{0.7442}{+0.0061} & \bestvalgain{20.72}{+1.64} & \bestvalgain{0.6945}{+0.0061} & \bestvalgain{22.22}{+1.09} & \bestvalgain{0.7362}{+0.0064} 
& \bestvalgain{28.29}{+0.80} & \bestvalgain{0.8592}{+0.0265} & \bestvalgain{23.83}{+1.31} & \bestvalgain{0.8020}{+0.0045} & \bestvalgain{13.78}{+0.20} & \valgain{0.5858}{+0.0075} & \bestvalgain{21.97}{+0.77} & \bestvalgain{0.7490}{+0.0128} \\
\midrule

\multirow[c]{4}{*}[-14pt]{\bfseries AdaIR} 
& Baseline & 22.92 & 0.7616 & 21.26 & 0.7378 & 18.58 & 0.6870 & 20.92 & 0.7288
& 27.71 & 0.8433 & 22.53 & 0.7981 & 13.93 & 0.5885 & 21.39 & 0.7433 \\[4pt]
& +StyleSSP & \valgain{18.89}{-4.03} & \valgain{0.6735}{-0.0881} & \valgain{18.59}{-2.67} & \valgain{0.6395}{-0.0983} & \valgain{17.34}{-1.24} & \valgain{0.6444}{-0.0426} & \valgain{18.27}{-2.65} & \valgain{0.6525}{-0.0763} 
& \valgain{16.27}{-11.44} & \valgain{0.6626}{-0.1807} & \valgain{18.19}{-4.34} & \valgain{0.5670}{-0.2311} & \valgain{13.02}{-0.91} & \bestvalgain{0.6626}{+0.0741} & \valgain{15.83}{-5.56} & \valgain{0.6510}{-0.0923} \\[4pt]
& +WTNet & \valgain{23.01}{+0.09} & \valgain{0.7422}{-0.0331} & \valgain{21.90}{+0.64} & \valgain{0.7257}{-0.0121} & \valgain{19.23}{+0.65} & \valgain{0.6513}{-0.0357} & \valgain{21.38}{+0.46} & \valgain{0.7064}{-0.0224} 
& \valgain{26.63}{-1.08} & \valgain{0.8450}{+0.0017} & \valgain{22.47}{-0.06} & \valgain{0.7770}{-0.0211} & \valgain{13.63}{-0.30} & \valgain{0.5879}{-0.0006} & \valgain{20.91}{-0.48} & \valgain{0.7366}{-0.0067} \\[4pt]
& +DDTNet & \bestvalgain{24.17}{+1.25} & \bestvalgain{0.7753}{+0.0137} & \bestvalgain{22.49}{+1.23} & \bestvalgain{0.7490}{+0.0112} & \bestvalgain{20.42}{+1.84} & \bestvalgain{0.7010}{+0.0140} & \bestvalgain{22.36}{+1.44} & \bestvalgain{0.7416}{+0.0128} 
& \bestvalgain{27.84}{+0.13} & \bestvalgain{0.8466}{+0.0033} & \bestvalgain{22.73}{+0.20} & \bestvalgain{0.7997}{+0.0016} & \bestvalgain{13.96}{+0.06} & \valgain{0.5955}{+0.0070} & \bestvalgain{21.51}{+0.12} & \bestvalgain{0.7472}{+0.0039} \\
\midrule

\multirow[c]{4}{*}[-14pt]{\bfseries DFPIR} 
& Baseline & 22.95 & 0.7525 & 21.12 & 0.7169 & 20.09 & 0.6752 & 21.19 & 0.7099
& 25.20 & 0.7232 & 22.18 & 0.7785 & 12.24 & 0.5146 & 19.87 & 0.6721 \\[4pt]
& +StyleSSP & \valgain{17.03}{-5.92} & \valgain{0.6768}{-0.0757} & \valgain{18.05}{-3.07} & \valgain{0.6479}{-0.0690} & \valgain{15.77}{-4.32} & \valgain{0.6363}{-0.0389} & \valgain{16.88}{-4.31} & \valgain{0.6509}{-0.0590} 
& \valgain{16.49}{-8.71} & \valgain{0.6403}{-0.0829} & \valgain{17.73}{-4.45} & \valgain{0.6880}{-0.0905} & \valgain{12.94}{+0.70} & \valgain{0.4811}{-0.0335} & \valgain{15.72}{-4.15} & \valgain{0.6031}{-0.0690} \\[4pt]
& +WTNet & \valgain{23.14}{+0.19} & \valgain{0.7539}{+0.0014} & \bestvalgain{21.96}{+0.84} & \bestvalgain{0.7234}{+0.0065} & \valgain{19.13}{-0.96} & \valgain{0.6428}{-0.0324} & \valgain{21.16}{-0.03} & \valgain{0.7001}{-0.0098} 
& \valgain{25.05}{-0.15} & \bestvalgain{0.8113}{+0.0811} & \valgain{22.04}{-0.14} & \valgain{0.7562}{-0.0223} & \valgain{13.25}{+1.01} & \valgain{0.5109}{-0.0037} & \valgain{20.11}{+0.24} & \valgain{0.6928}{+0.0207} \\[4pt]
& +DDTNet & \bestvalgain{24.04}{+1.09} & \bestvalgain{0.7645}{+0.0120} & \valgain{21.80}{+0.68} & \valgain{0.7194}{+0.0025} & \bestvalgain{21.16}{+1.07} & \bestvalgain{0.6805}{+0.0053} & \bestvalgain{22.13}{+0.94} & \bestvalgain{0.7160}{+0.0061} 
& \bestvalgain{27.14}{+1.94} & \valgain{0.7958}{+0.0726} & \bestvalgain{22.21}{+0.03} & \bestvalgain{0.7800}{+0.0015} & \bestvalgain{13.40}{+1.16} & \bestvalgain{0.5431}{+0.0285} & \bestvalgain{20.91}{+1.04} & \bestvalgain{0.7064}{+0.0343} \\
\midrule

\multicolumn{2}{c|}{\bfseries DDTNet Avg Gain} 
& \bfseries +1.07 & \bfseries +0.0112 
& \bfseries +0.89 & \bfseries +0.0066 
& \bfseries +1.52 & \bfseries +0.0084 
& \bfseries +1.16 & \bfseries +0.0084 
& \bfseries +0.96 & \bfseries +0.0341 
& \bfseries +0.62 & \bfseries +0.0092 
& \bfseries +0.47 & \bfseries +0.0143 
& \bfseries +0.68 & \bfseries +0.0170 \\
\bottomrule
\end{tabular}%
}
\endgroup
\end{table*}

\begin{figure*}[t!]
\centering
\includegraphics[width=0.99\textwidth]{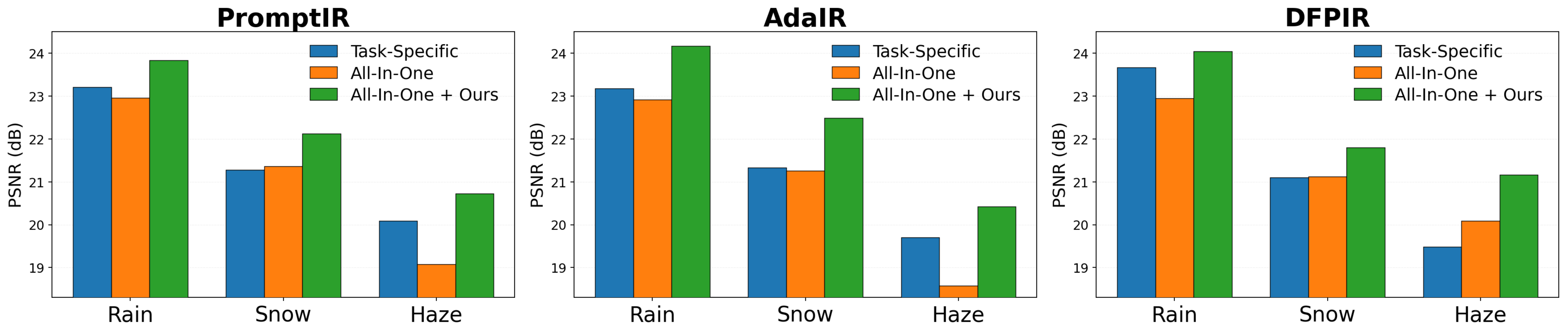}
\vspace{-0in}
\caption{
\textbf{Quantitative comparison among three training configurations: 1) task-specific training, 2)
all-in-one training, and 3) all-in-one training enhanced with
DDTNet.}
All-in-one methods (PromptIR, AdaIR, and DFPIR) often underperform their task-specific versions.
In contrast, DDTNet enhances the performance of all-in-one models, allowing them to surpass task-specific performance.
}
\label{fig:Training_Diff}
\vspace{-0.13in}
\end{figure*}

\begin{figure*}[t!]
    \centering
    \includegraphics[width=\textwidth]{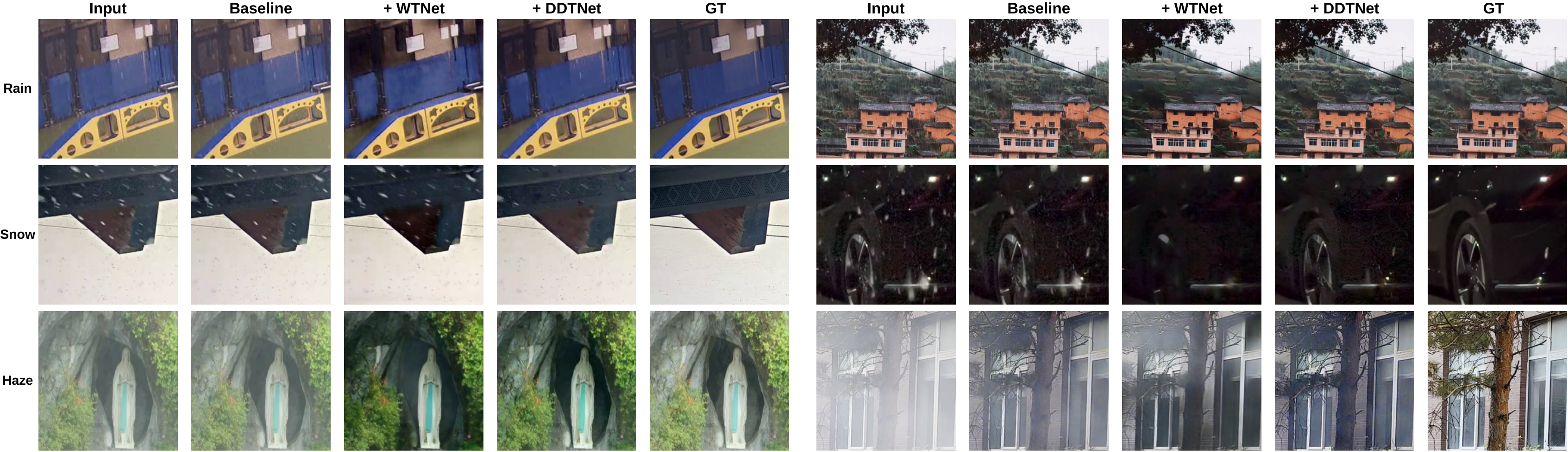}
    \caption{
    \textbf{Qualitative comparison of PromptIR on WeatherStream (left) and WeatherBench (right)} between its baseline and its WTNet- and DDTNet-enhanced versions under rain, snow, and haze conditions.}
    \vspace{-0.13in}
    \label{fig:promptir_res}
    
    \vspace{0.3in} 
    
    \includegraphics[width=\textwidth]{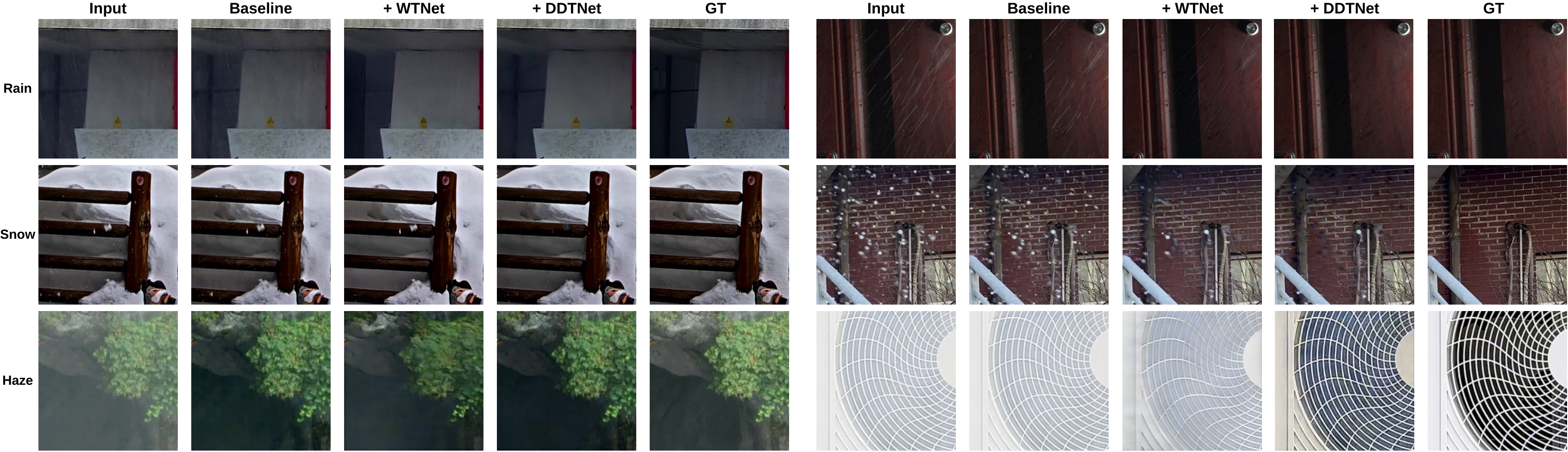}
    \caption{
    \textbf{Qualitative comparison of DFPIR on WeatherStream (left) and AdaIR on WeatherBench (right)} between their baselines and their WTNet- and DDTNet-enhanced versions under rain, snow, and haze conditions.
    }
    \vspace{-0.13in}
    \label{fig:adair_dfpir_result}
    \vspace{-0.in}
\end{figure*}

\subsection{Quantitative Comparison}
{\color{black}
Table~\ref{table:restoration_two} presents a comprehensive quantitative comparison of different test-time domain adaptation (TTDA) methods integrated into three state-of-the-art (SOTA) all-in-one image restoration models: PromptIR~\cite{potlapalli2023promptir}, AdaIR~\cite{cui2025adair}, and DFPIR~\cite{tian2025degradation}. In the table, ``Baseline'' denotes these SOTA models trained without any adaptation, while the subsequent rows show the results of applying StyleSSP~\cite{xu2025stylessp}, WTNet~\cite{Huang_2025_BMVC}, and our proposed DDTNet to these baselines for TTDA.

Compared to existing TTDA approaches, DDTNet demonstrates superior robustness and adaptability. Specifically, StyleSSP suffers from severe performance degradation on the WeatherBench dataset (e.g., PSNR drops of up to $11.44$~dB on AdaIR deraining) and fails to generalize effectively across varying degradation types. Meanwhile, WTNet provides only marginal gains or even negative impacts in several scenarios (e.g., a $1.88$~dB drop on PromptIR deraining in WeatherBench). In contrast, DDTNet consistently and significantly boosts performance across all baseline models and datasets.

On WeatherStream~\cite{zhang2023weatherstream}, DDTNet yields impressive performance gains, boosting the average PSNR by $1.09$~dB for PromptIR, $1.44$~dB for AdaIR, and $0.94$~dB for DFPIR. When investigated by task, DDTNet achieves overall average PSNR improvements of $1.07$~dB for deraining, $0.89$~dB for desnowing, and $1.52$~dB for dehazing.
On WeatherBench~\cite{guan2025weatherbench}, DDTNet likewise achieves notable gains, improving the average PSNR by $0.77$~dB, $0.12$~dB, and $1.14$~dB for PromptIR, AdaIR, and DFPIR, respectively. For each specific task, DDTNet delivers average PSNR improvements of $0.96$~dB for deraining, $0.62$~dB for desnowing, and $0.47$~dB for dehazing.

Furthermore, Figure~\ref{fig:Training_Diff} compares the three restoration models under three training configurations: 1) task-specific training, 2) all-in-one training, and 3) all-in-one training enhanced with DDTNet. The results confirm that all-in-one training typically suffers from inter-task interference, yielding suboptimal results compared to task-specific training. In contrast,  DDTNet consistently improves performance across all models and tasks, even outperforming their task-specific counterparts. Overall, these results demonstrate that DDTNet not only improves restoration performance on unseen target domains but also effectively mitigates the inherent limitations of conventional all-in-one training schemes.
}

We also compare DDTNet with the general-purpose domain adaptation approach for this task, Noise-DA~\cite{liao2024denoising}. 
To ensure a fair comparison, we adopt the UNet backbone used in Noise-DA as the baseline restoration model. We also re-implement Noise-DA with its official settings and train it on our dataset under the all-in-one training scheme.
As reported in Table~\ref{table:unet-ws}, Noise-DA exhibits limited generalization: while it yields gains on deraining and desnowing, it causes a performance drop on dehazing, reflecting its lack of adaptability to diverse degradations in an all-in-one setting.
In contrast, DDTNet disentangles heterogeneous degradation patterns to generate domain-adaptive pairs, consistently boosting the restoration performance across all tasks and achieving a substantial average PSNR gain of $5.67$ dB over the baseline.

\begin{figure}[t!]
\centering
\includegraphics[width=\linewidth]{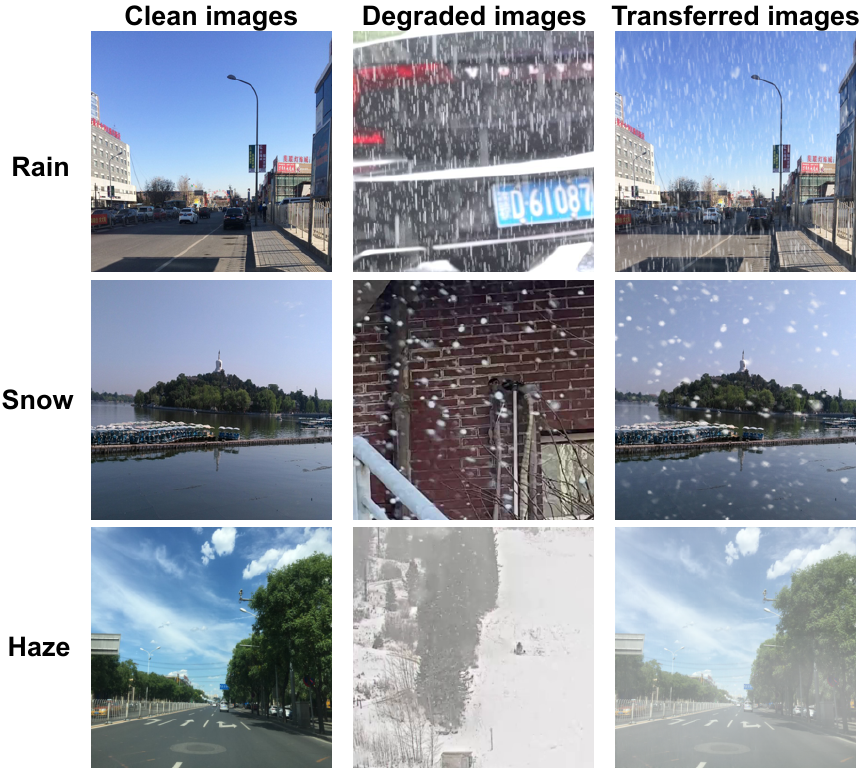}
\caption{
\textbf{Examples of degradation-transferred images.} Degradation patterns in the target-domain images from WeatherStream and WeatherBench are transferred to clean source-domain images from RESIDE.
}
\label{fig:transferred_img}
\vspace{-0.13in}
\end{figure}

\begin{table}[t]
\centering
\begingroup
\setlength{\tabcolsep}{4.5pt}
\renewcommand{\arraystretch}{1.12}

\caption{\textbf{Quantitative comparison between DDTNet and Noise-DA on WeatherStream}, with the restoration backbone employed by Noise-DA as the baseline.}
\vspace{-0in}
\label{table:unet-ws}

\newcommand{\gaincolor}{black!65}     
\newcommand{\gainfontsize}{\footnotesize} 

\newcommand{\valgain}[2]{%
  \begin{tabular}{@{}c@{}}%
    #1 \\[-3pt]%
    {\gainfontsize\color{\gaincolor}(#2)}%
  \end{tabular}%
}
\newcommand{\valgainbf}[2]{%
  \begin{tabular}{@{}c@{}}%
    \bfseries #1 \\[-3pt]%
    {\gainfontsize\bfseries\color{\gaincolor}(#2)}%
  \end{tabular}%
}

\resizebox{\linewidth}{!}{%
\begin{tabular}{@{} c c | c c | c c | c c | c c @{}}
\toprule
& & \multicolumn{2}{c|}{Rain} & \multicolumn{2}{c|}{Snow} & \multicolumn{2}{c|}{Haze} & \multicolumn{2}{c}{Average} \\
\multicolumn{2}{c|}{Method} & PSNR~$\uparrow$ & SSIM~$\uparrow$ & PSNR~$\uparrow$ & SSIM~$\uparrow$ & PSNR~$\uparrow$ & SSIM~$\uparrow$ & PSNR~$\uparrow$ & SSIM~$\uparrow$ \\
\midrule
\multicolumn{2}{c|}{Baseline}
  & 16.23 & 0.496 & 15.32 & 0.484 & 18.06 & 0.561 & 16.74 & 0.514 \\
\midrule
\multicolumn{2}{c|}{+Noise-DA}
  & \valgain{19.11}{+2.88} & \valgain{0.596}{+0.100}
  & \valgain{17.24}{+1.92} & \valgain{0.577}{+0.093}
  & \valgain{16.93}{-1.13} & \valgain{0.583}{+0.022}
  & \valgain{17.76}{+1.02} & \valgain{0.586}{+0.072} \\
\multicolumn{2}{c|}{\rule{0pt}{2.6ex}\textbf{+DDTNet}}
  & \valgainbf{24.56}{+8.33} & \valgainbf{0.779}{+0.283}
  & \valgainbf{21.55}{+6.23} & \valgainbf{0.735}{+0.251}
  & \valgainbf{21.12}{+3.06} & \valgainbf{0.699}{+0.138}
  & \valgainbf{22.41}{+5.67} & \valgainbf{0.737}{+0.223} \\
\bottomrule
\end{tabular}}
\vspace{-0.13in}
\endgroup
\end{table}

\subsection{Qualitative Comparison}
We present qualitative comparisons of de-weathering results using three restoration models, including PromptIR in Figure~\ref{fig:promptir_res}, and AdaIR and DFPIR in Figure~\ref{fig:adair_dfpir_result}. In these visualizations, ``Baseline'' denotes the models trained without DDTNet, whereas ``+WTNet'' and ``+DDTNet'' indicate their WTNet- and DDTNet-enhanced counterparts, respectively.
The baseline models often struggle to generalize to diverse degradation patterns in the target domains. {\color{black} Although WTNet achieves promising results in degradation removal, it tends to compromises fine-grained content details, leading to noticeable image distortion.} In contrast, the DDTNet-enhanced models effectively remove complex degradations, producing visually superior and perceptually more realistic results.

Figure~\ref{fig:transferred_img} further shows examples of degradation-transferred images, where the degraded inputs are selected from WeatherStream~\cite{zhang2023weatherstream} and WeatherBench~\cite{guan2025weatherbench}, and the clean images are sampled from RESIDE. 
In each case, DDTNet faithfully replicates the degradation patterns of the degraded images while strictly preserving the scene content of the clean image. 
These results demonstrate that DDTNet can successfully generate domain-adaptive pairs that align with the target-domain degradation distribution, thereby enabling effective fine-tuning of restoration models and improving their performance during inference.
These visualizations highlight DDTNet’s strong cross-domain generalization capability in handling diverse real-world weather degradations.

\subsection{Ablation Studies}
To analyze the quality of the generated degradation-transferred images, we construct a test set of $900$ degraded–clean-mixed triplets $\{(I^d_i, I^c_i, I^{m}_i)\}_{i=1}^{900}$ on Rain100H/Rain100L~\cite{Yang_2017_CVPR},  Snow100K~\cite{liu2018desnownet}, and RESIDE~\cite{li2019benchmarking}, with 300 triplets per task, to synthesize rain, snowy, and hazy images. 
Based on this dataset, we provide a comprehensive component analysis of the proposed DDTNet.

\paragraph{Ablation Study of DDTNet}
Table~\ref{tab:ddm-ablation-left} presents an ablation study of the two DCA components within DDTNet: $\mathrm{DCA_S}$ and $\mathrm{DCA_C}$. 
We implement four variants of the network, including  1) Net1: a vanilla encoder–decoder architecture without DDM, serving as the baseline; 2) Net2: Net1 augmented with $\mathrm{DCA_S}$ alone in the DDM; 3) Net3: Net1 augmented with $\mathrm{DCA_C}$ alone in the DDM; and 4) Ours: the complete DDTNet with both $\mathrm{DCA_S}$ and $\mathrm{DCA_C}$. 
{\color{black}
The results indicate that both Net2 and Net3 enhance the average performance of the baseline (Net1), yielding average PSNR gains of $2.71$ dB and $0.85$ dB, respectively.
Our proposed DDTNet achieves the best overall performance among all variants.
Specifically, DDTNet improves the average PSNR of the baseline encoder-decoder model by $2.94$ dB across the three degradation removal tasks.
}

\begin{table}[t]
\centering
\caption{
\textbf{Component analysis of DDTNet} based on the average PSNR of degradation-transferred images across Rain100H/L, Snow100K, and RESIDE.
}
\vspace{-0in}
\label{tab:ddm-ablation-left}
\begingroup
\setlength{\tabcolsep}{8pt}%
\renewcommand{\arraystretch}{1.12}
\setlength{\abovecaptionskip}{2pt}%
\setlength{\belowcaptionskip}{2pt}%
\setlength{\aboverulesep}{0.4ex}%
\setlength{\belowrulesep}{0.4ex}%
\footnotesize 

\begin{tabular}{c | c | c c | c}
\toprule
 & \multirow{2}{*}{Enc–Dec} & \multicolumn{2}{c|}{DDM} & \multirow{2}{*}{PSNR} \\
\cline{3-4}
 &  & $\mathrm{DCA_S}$ & $\mathrm{DCA_C}$ &  \\
\midrule
Net1 & \ding{51} &  &  & 31.59 \\
Net2 & \ding{51} & \ding{51} &  & 34.30 \\
Net3 & \ding{51} &  & \ding{51} & 32.44 \\
Ours & \ding{51} & \ding{51} & \ding{51} & \textbf{34.53} \\
\bottomrule
\end{tabular}

\endgroup
\end{table}

\begin{table}[t!]
\centering
\caption{
\textbf{Component analysis of Degradation-Coupled Attention (DCA)} based on the average PSNR of degradation-transferred images across Rain100H/L, Snow100K, and RESIDE.}
\vspace{-0in}
\label{tab:ddm-ablation-right}
\begingroup
\setlength{\tabcolsep}{8pt}
\renewcommand{\arraystretch}{1.25}
\resizebox{\linewidth}{!}{
\begin{tabular}{c | c | c | c c | c}
\toprule
 & \multirow{2}{*}{Enc--Dec} & \multirow{1}{*}{Degr.-Sensing} & \multicolumn{2}{c|}{Attention Layers} & \multirow{2}{*}{PSNR} \\
\cline{4-5}
 &  & Tokens & Attn--L1 & Attn--L2 &  \\
\midrule
Net1 & \ding{51} &  &  &  & 31.59 \\
Net4 & \ding{51} &  & \ding{51} & \ding{51} & 33.00 \\
Net5 & \ding{51} & \ding{51} &  & \ding{51} & 33.35 \\
Ours & \ding{51} & \ding{51} & \ding{51} & \ding{51} & \textbf{34.53} \\
\bottomrule
\end{tabular}}
\endgroup
\vspace{-0.13in}
\end{table}

\paragraph{Ablation Study of DCA}
Table~\ref{tab:ddm-ablation-right} analyzes the effectiveness and contributions of the proposed Degradation-Coupled Attention (DCA). 
DCA is composed of three internal components: degradation-sensing tokens, Attention Layer 1, and Attention Layer 2. 
We compare four module configurations, including
1) Net1: a pure encoder–decoder architecture without DDM, identical to the baseline in Table~\ref{tab:ddm-ablation-left};
2) Net4: Net1 augmented with Attention Layers $1$ and $2$ in DCA, but without degradation-sensing tokens. In this case, the attention layers directly perform self-attention on the input features, functioning as a standard data-driven mechanism;
3) Net5: Net1 augmented with degradation-sensing tokens and Attention Layer $2$, where the tokens are directly fed into the second attention layer without applying the designed coupling mechanism; and
4) Ours: the full DDTNet with all DCA components.
The comparison reveals that both degradation-sensing tokens and the coupled attention layers individually improve the baseline. 
Crucially, the complete integration of all three components significantly outperforms the other variants. 
This verifies the effectiveness of the proposed DCA in facilitating robust degradation transfer.

\begin{figure*}[t!]
\centering
\includegraphics[width=0.99\textwidth]{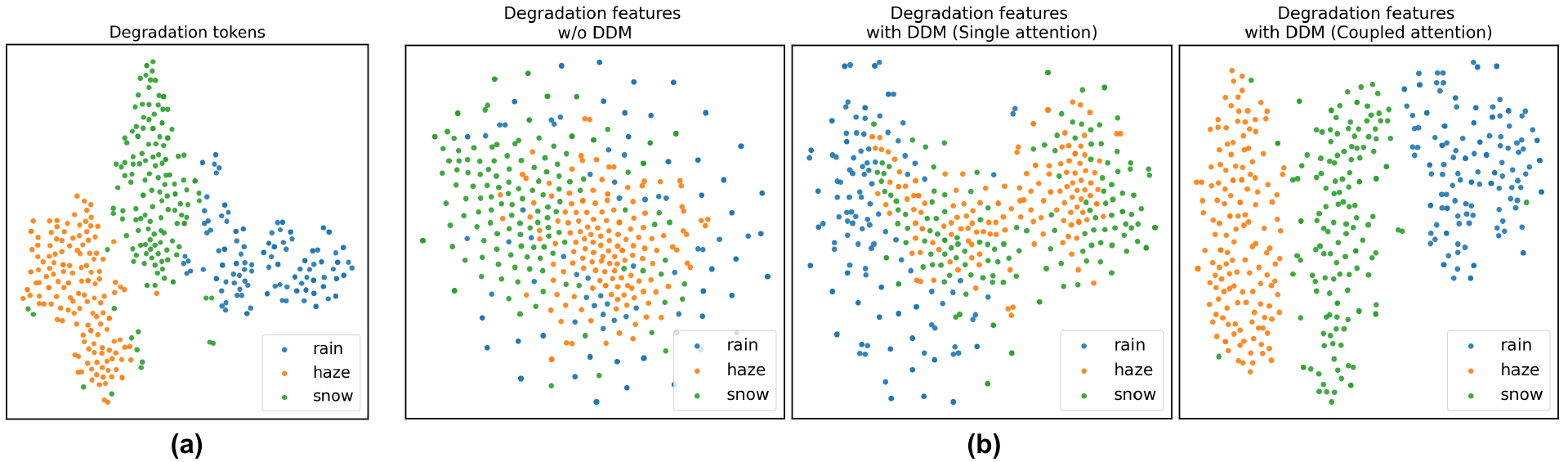}
\vspace{-0in}
\caption{
\textbf{Visualization of the learned degradation tokens and degradation features.}
%
(a) t-SNE visualization of the learned degradation tokens $\tilde{T}$. (b) t-SNE analysis of the degradation features at different stages: (left) encoder output $\tilde{X}$ without DDM, (middle) DCA without Attention Layer 1, and (right) the DDM output of the full DDTNet.
}
\label{fig:tsne}
\vspace{-0.13in}
\end{figure*}

\begin{table}[t!]
\small
\centering
\setlength{\tabcolsep}{1mm} 
\renewcommand{\arraystretch}{1.3} 
\caption{
\textbf{Ablation study of the degradation-sensing token count ($M$)} on degradation transfer quality.
}
\vspace{-0in}
\resizebox{\linewidth}{!}{
\begin{tabular}{c|ccccc}
\noalign{\hrule height 1.0pt}
Number of Tokens ($M$) & 32 & 64 & 128 & 256 & 512 \\
\noalign{\hrule height 0.5pt}
Average PSNR (dB) & 32.78 & 32.93 & 33.46 & \bf 34.49 & 34.01 \\
\noalign{\hrule height 1.0pt}
\end{tabular}
}
\label{tab:ablation_tokens}
\vspace{-0.13in}
\end{table}

\paragraph{Impact of Degradation-Sensing Token Count}
{\color{black}
To determine the optimal representational capacity for capturing complex weather conditions, we investigate the impact of the number of degradation-sensing tokens. Table~\ref{tab:ablation_tokens} reports the degradation transfer quality in terms of average PSNR across different token counts $M \in \{32, 64, 128, 256, 512\}$. As the results indicate, increasing the number of tokens from $32$ to $256$ yields a consistent performance improvement, with the average PSNR rising from $32.78$ dB to a peak of $34.49$ dB. This suggests that a longer token sequence provides the network with a larger capacity to extract richer and more fine-grained degradation representations. 
However, further increasing $M$ to $512$ results in a slight performance decline to $34.01$ dB, likely due to the potential informational redundancy among these tokens and the risk of overfitting on the degradation representations. 
Consequently, we empirically set $M = 256$ as the default configuration for DDTNet to achieve the best trade-off between representational capacity and restoration performance.
}


\paragraph{Analysis of the Degradation Tokens and Features} Figure~\ref{fig:tsne}(a) presents a t-SNE visualization of the learned degradation tokens $\tilde{T}$, obtained by randomly sampling 150 images per task (deraining, desnowing, and dehazing).
These learned tokens exhibit clear clustering based on their weather types: rain samples (blue) are grouped in the middle-right, haze samples (orange) are in the lower-left, and snow samples (green) are near the center.
This clear separation across weather types, combined with strong intra-class consistency, confirms that the degradation tokens $\tilde{T}$ effectively capture and disentangle degradation-specific features.

%
Furthermore, Figure~\ref{fig:tsne}(b) presents the t-SNE plots of the output features produced by three module configurations.
Its left plot corresponds to the model without the DDM (i.e., Net1 in Table~\ref{tab:ddm-ablation-right}), where the encoder output features $\tilde{X}$ for rain, snow, and haze are heavily entangled. 
The middle plot corresponds to Net5 in Table~\ref{tab:ddm-ablation-right}, which retains only Attention Layer 2. The output features $\tilde{X}$ still exhibit substantial overlap, though rain features can be separated from other weather types. 
As shown in the right plot, the full DDM produces distinct, weather-specific clusters.
These results show that the DDM effectively disentangles degradation patterns and yields discriminative features, enabling robust degradation transfer in all-in-one restoration.

\section{Conclusion}

This paper introduces the Degradation Disentanglement and Transfer Network (DDTNet), a novel approach designed for domain adaptation in all-in-one image de-weathering.
To address the common challenge where paired degraded–clean images are typically unavailable during target-domain inference, DDTNet effectively transfers degradation patterns from target-domain degraded images onto source-domain clean images. 
This process generates domain-adaptive pairs for test-time fine-tuning. 
%
%
The technical cornerstone of DDTNet is the Degradation-Coupled Attention (DCA) mechanism, which integrates learnable degradation-sensing tokens to extract degradation cues and then distills these degradation features through a two-stage attention mechanism. 
%
%
%
Extensive experimental results demonstrate that DDTNet substantially improves state-of-the-art all-in-one restoration models across real-world deraining, dehazing, and desnowing benchmarks, validating its effectiveness for robust cross-domain adaptation.
Furthermore, while this study focuses on adverse-weather restoration, DDTNet’s generic encoder–decoder design and attention-based degradation transferability make it applicable to other restoration tasks, where degraded–clean–mixed triplets are available, providing a promising foundation for future research.


\bibliographystyle{IEEEtran}
\bibliography{main}

\vfill

\end{document}